%% file: paper_iv2019.tex
\documentclass[letterpaper, 10pt, conference]{ieeeconf} %% FINAL
\IEEEoverridecommandlockouts    % This command is only needed if 
\usepackage[utf8]{inputenc}
\usepackage[noadjust]{cite} % from IEEEtran documentation: 
\usepackage{microtype}
\usepackage{amsmath,amssymb,amsfonts,bm}
\usepackage{multirow}
\usepackage{hhline}
\usepackage{siunitx}
\usepackage{threeparttable}
\usepackage{url}
\usepackage[dvipsnames ]{xcolor}
\usepackage{prettyref}
\usepackage{blindtext}
\usepackage{placeins}
\usepackage{booktabs}

\usepackage{tikz,pgfplots}

\usepackage{hyperref}

\pgfplotsset{compat=1.13}
\usetikzlibrary{
    external,
    calc,
    arrows,
    shapes,
    positioning,
    through,
    patterns,
}
\hypersetup{hidelinks}

\newcommand{\valunit}[2]{\ensuremath{#1\,\text{#2}}} % value with unit and small space
\newcommand{\ul}[1]{\underline{#1}}
\newcommand{\htodo}[1]{} % hidden todo

\newcommand{\htodos}[1]{} % hidden todos

\newrefformat{eq}{\textup{(\ref{#1})}}
\newrefformat{sec}{\mbox{Sec.\,\ref{#1}}}
\newrefformat{subsec}{\mbox{Sec.\,\ref{#1}}}
\newrefformat{tab}{Tab.\,\ref{#1}}
\newrefformat{fig}{Fig.\,\ref{#1}}
\newrefformat{alg}{Algorithm~\ref{#1}}
\let\pref\prettyref
\newcommand{\ie}{i.\,e.\xspace}
\newcommand{\eg}{e.\,g.\xspace}

\newcommand{\cf}{c.\,f.\xspace}

\newcommand{\Transp}{\mathrm{T}}
\newcommand{\mylvec}[1]{\begin{bmatrix}#1\end{bmatrix}^\Transp}
\renewcommand{\vec}[1]{\ensuremath{\boldsymbol{#1}}}    % vectors (bold)
\newcommand{\percent}{\%}
\newcolumntype{L}[1]{>{\raggedright\let\newline\\\arraybackslash\hspace{0pt}}m{#1}}
\newcolumntype{C}[1]{>{\centering\let\newline\\\arraybackslash\hspace{0pt}}m{#1}}
\newcolumntype{R}[1]{>{\raggedleft\let\newline\\\arraybackslash\hspace{0pt}}m{#1}}

\begin{document}

% space saving (figures)
\addtolength{\textfloatsep}{-10pt} % between float and text (recommendation -10pt)
\addtolength{\abovecaptionskip}{-5pt} % above (figure) caption (recommendation -5pt)
\addtolength{\floatsep}{-5pt} % between two floats (recommendation -5pt)
%\addtolength{\intextsep}{-5pt} % between float and text (for [h] floats)

% space saving (equations)
\addtolength{\abovedisplayskip}{-1pt} % space above equations
\addtolength{\abovedisplayshortskip}{-1pt} % space above short equation
%\addtolength{\belowdisplayskip}{-1pt} % space below equations (not recommended!)
%\addtolength{\belowdisplayshortskip}{-1pt} % space below short equations (not recommended!)
    
    \htodos{
        \begin{itemize}
            \item reposition \textit{newpage} command for equalization of last page
        \end{itemize}
    }
    
    \title{
        \LARGE\bf
        Generating Compact Geometric Track-Maps\\
        for Train Positioning Applications
    }    
    
    \newcommand{\mrmr}{\textsuperscript{*}}
    \author{
        Hanno Winter\mrmr,
        Stefan Luthardt\mrmr, 
        Volker Willert\mrmr, 
        Jürgen Adamy\mrmr%
        \thanks{
            \mrmr{}Control Methods and Robotics Laboratory, TU Darmstadt,
            Germany\newline 
            \textit{\{hanno.winter,stefan.luthardt,vwillert,adamy\}@rmr.tu-darmstadt.de}
        }%
    }
    
    \maketitle
    
    \input{content/my_content}

    %\printlen[2][cm]{\columnwidth}    
    %\newpage
    
%     \IEEEtriggeratref{7}
%     \IEEEtriggercmd{\enlargethispage{-5.35in}}

    % \bibliographystyle{IEEEtran_v1_12}
    \bibliographystyle{IEEEtran_v1_14}
    \bibliography{IEEEabrv,my_references}
    %\nocite{*}

\end{document}

%% file: content/my_content.tex
\begin{abstract} % _____________________________________________________________
% A “stand alone” condensed version of the article
% • No more than 250 words; written in the past tense
% • Uses keywords and index terms
% • Answer the following questions:
%     • Why you did
%     • What you did
%     • How the results were useful, important & move the field forward
%     • Why they’re useful & important move the field forward
In this paper, we present a method to generate compact geometric track-maps for
train-borne localization applications. Therefore, we first give a brief overview
on the purpose of track maps in train-positioning applications. It
becomes apparent that there are hardly any adequate methods to generate suitable
geometric track-maps. This is why we present a novel map generation procedure.
It uses an optimization formulation to find the continuous sequence of track
geometries that fits the available measurement data best. The optimization is
initialized with the results from a localization filter
\cite{my_references:WinterWillertEtAl2018IncreasingAccuracyTrain} developed in
our previous work. The localization filter also provides the required
information for shape identification and measurement association. The presented
approach will be evaluated on simulated data as well as on real measurements.
\end{abstract}

% \begin{IEEEkeywords}  % ________________________________________________________       
%      Rail transportation, 
%      Sensor fusion, 
%      Simultaneous localization and mapping, 
%      Systems modeling, 
%      Recursive estimation, 
%      Kalman filters 
% \end{IEEEkeywords}

\section{Introduction} % _______________________________________________________
% • A description of the problem you researched
% • It should move step by step through, should be written in present tense:
%     • Generally known information about the topic
%     • Prior studies' historical context to your research
%     • Your hypothesis and an overview of the results
%     • How the article is organized
% • The introduction should not be:
%     • Too broad or vague
%     • More then 2 pages

For safety reasons, trains currently operate strictly signal-based.
Therefore, each track is divided in multiple block sections delimited by
signals. The signals are coordinated by a central safety logic which guarantees
that only one train can occupy a block section at the same time. The necessary
train-position information is gathered by sensors installed at the tracks.
Although this system has proven to be safe and reliable it suffers
either from high costs for the huge amount of sensors, or a low track capacity
due to longer block sections
\cite{my_references:TheegVlasenko2009Railwaysignalling}. To overcome this
undesirable trade-off in the near future, trains have to become intelligent
vehicles which are able to localize themselves continuously without any
track-side installations.

The challenge when developing such a train-borne localization system is to
fulfill the high demands in terms of \ul{r}eliability, \ul{a}vailability,
\ul{m}aintainability, and \ul{s}afety (RAMS) in the sense of EN 50126
\cite{my_references:en50126}.
Although the development of train-borne localization systems has gained of
interest in recent years, there is currently no sensor configuration available
fulfilling all demands
\cite{my_references:OteguiBahilloEtAl2017SurveyTrainPositioning}.

In this paper, we want to focus on the purpose of digital track-maps in
train-borne localization systems and how they can help to fulfill the RAMS
demands in the near future. Therefore, we will investigate how track maps are
utilized to improve the positioning accuracy, availability and integrity of
train-borne localization systems. After that, we will give a brief overview over
the methods commonly used to generate track maps. From this overview it will
become apparent that there are hardly any adequate methods to generate suitable
track maps for the purposes described above. Therefore, we will present a new
approach to generate compact geometric track-maps based on the results of a
localization filter we presented in our previous work
\cite{my_references:WinterWillertEtAl2018IncreasingAccuracyTrain} motivated by
\cite{my_references:SchreierWillertEtAl2014Gridmappingdynamic}.

\section{Track Maps in Train-Borne Localization} % _____________________________
\label{sec:track_maps_for_localization}

We start with a brief overview on the different types of track maps and the
methods used to generate them. Afterwards, we briefly discuss the shortcomings
of these methods which motivated us to come up with a novel approach to
generate compact geometric track-maps.

\subsection{Map Types}
\label{sec:track_maps_types}

There are three different categories of track maps used for train-borne
localization:

\subsubsection{Topological Track-Maps}
This is the most basic track-map type. It only stores the topology and mileage
of the railway network. This is sufficient due to the fact that the position $p$
of a train can unambiguously be defined in railway coordinates by $p =
\{t,\,s\}$, where $t$ represents a unique track ID and $s$ being a
continuous track-length parameter. These maps are widely used in the railway
system today since an additional absolute position information is not needed
for its safe operation.

Theoretically, it is possible to realize a train-borne localization system with
these maps. Therefore, it has to be assumed that the start point and the
pre-set route of a specific train is known. 
Then a train can localize itself by measuring its traveled distance relative to
its start point and thereby determining its position on this pre-set route
\cite{my_references:SchneiderTroelsen2000Introducingdigitalmap}. 
Unfortunately, the pre-set route is normally not known on the train itself.
This makes the localization result ambiguous: After a switch the position can
no longer be clearly determined. To solve this ambiguity, maps holding
additional information have been introduced as described next.

\subsubsection{Topographic and Geometric Track-Maps}
Compared to the topological track-maps described before, topographic track-maps
additionally store the track-course in absolute coordinates. Furthermore, if
they hold track-characteristic information like the specific track element type
(straight, circular arc or transitional arc), orientation, curvature, or
something similar, they can be additionally named geometric track-maps. The
additional information stored in these maps allows to apply different
map-matching techniques with more track-selective localization approaches
compared to topological track-maps.

The map-matching approaches vary depending on the used sensor configuration.
Many approaches utilize global navigation satellite system (GNSS) data and
inertial measurement unit (IMU) data together with course and curvature
information from a track map to realize track-selective map-matching
\cite{my_references:Saab2000mapmatchingapproachb,
my_references:LueddeckeRahmig2011EvaluatingmultipleGNSS,
my_references:BroquetasComeronEtAl2012Trackdetectionrailway,
my_references:CrespilloHeirichEtAl2014BayesianGNSSIMUtight,
my_references:RothBaaschEtAl2018MapSupportedPositioning}. However, the
additional map information can not only be used to improve the localization
accuracy. It can also be used to increase the availability and integrity of the
localization system itself
\cite{my_references:NeriPalmaEtAl2013TrackconstrainedPVT,
my_references:YuEun2017Sensorattackdetection,
my_references:JinCaiEtAl2018DTMaidedadaptive}.

\subsubsection{Feature Track-Maps}
This type of track-map also stores information on features or landmarks along
the track. Features directly used for train-borne localization are for example
ferromagnetic inhomogeneities of the rails
\cite{my_references:SpindlerLauer2018HighAccuracyEstimation} or characteristic
distortions of the earth magnetic-field along the railway track
\cite{my_references:SieblerHeirichEtAl2018TrainLocalizationParticle}. Other
features may be characteristic infrastructure elements like bridges, tunnels or
stations, as suggested in
\cite{my_references:GerlachHoerste2009precisedigitalmap} which can help to
increase the accuracy and availability of GNSS positioning results.

\subsection{Generation Methods}
\label{sec:track_maps_generation}

There are basically four main approaches to create digital track-maps
\cite{my_references:GerlachHoerste2009precisedigitalmap}:
\begin{itemize}
    \item Extraction from existing site plans, available as paper drawings,
    Computer Aided Design (CAD) plans, or Geographic Information System (GIS)
    databases,
    \item direct surveying of the tracks, \eg by GNSS measurements or the
    application of tachymetry,
    \item analysis of orthophotos, or
    \item the application of simultaneous localization and mapping (SLAM)
    methods.
\end{itemize}
Although track maps are indispensable for train-borne localization, it is often
not described in detail how the necessary maps are created. The probably most
commonly used maps consist of previously recorded position data-points which
are available from the localization sensors. If additional geometric track
information is needed, it is mostly referred to the possibility to extract this
data from existing site plans. To our knowledge there are currently two
simultaneous localization and mapping (SLAM) approaches available which are
especially designed for railway vehicles
\cite{my_references:HeirichRobertsonStrang2013RailSLAMLocalizationrail,
my_references:HasbergHenselEtAl2012Simultaneouslocalizationmapping}. Both
methods create data-point based track maps. In the following, we present a new
method that is not based on data points but on a concatenation of geometric
entities.

\subsection{Conclusions for Track-Maps and their
Generation}\label{sec:maps_discussion}

Based on the explanations in \pref{sec:track_maps_types} it becomes
obvious how important track-maps are for train-borne localization. They help to
increase the positioning accuracy, availability and integrity of the
localization system. Thus, track maps act like an additional passive sensor
helping to meet the RAMS requirements. This especially applies to geometric and
feature track-maps. However, it should also be stressed that an inaccurate track
map can also pose a single point of failure in the overall localization process.

The usability of map information for localization purposes is largely
influenced by the map representation and the map quality.
% which in turn depends on the map generation process.
%Consequently it makes sense to discuss some of the shortcomings of the methods
%used to generate those maps (\cf \pref{sec:track_maps_generation}).
For a map to be suitable for train-borne localization it has to fulfill at least
two basic requirements\footnote{Some further conclusions on the requirements
for digital track-maps as well as some modeling schemes can be found in
\cite{my_references:BikkerKlingeEtAl1998ConceptsIntelligentRoute,
my_references:BoehringerGeistler2006Locationrailwaytraffic,
my_references:GerlachHoerste2009precisedigitalmap}.}:
\begin{itemize}
    \item Track-length accuracy: It is essential to consistently
    assign all stored information with respect to the track-length $s$ since all
    localization algorithms somehow rely on this assignment.
    \item Compactness: All information must be accessible in a
    computationally efficient way, as the map is often directly used in the
    localization algorithm itself, which has to run in real-time. Furthermore,
    it is advantageous if the map consumes as little memory as possible in
    order to be easily transferable.
\end{itemize}

All current generation methods directly utilizing measurement data store the map
in a data-point format. Between neighboring data points interpolation techniques
are applied. To avoid large interpolation errors the tracks are normally densely
sampled, \ie with a sample distance between \valunit{1}{m} and \valunit{30}{m}.
Due to the necessary interpolation, such maps are not computationally efficient
and the resulting map representation is neither easy accessible nor  memory
saving. Thus, these maps are not optimal in the sense of the compactness
requirement mentioned above
\cite{my_references:LiuCaiEtAl2013Generatingelectronictrack}. A more suitable
track map representation would be a direct description of the geometric
properties of each track element in a list. This would result in geometric
track-maps easily fulfilling the compactness requirement. Those maps may be
extracted from existing site plans. However, these site plans can differ
significantly from the real track situation
\cite{my_references:LiPuEtAl2019MethodAutomaticallyRereating}. Two possible ways
to create compact geometric track-maps based on measurement data are presented in
\cite{my_references:TaoCaiEtAl2017Digitaltrackmap,
my_references:LiPuEtAl2019MethodAutomaticallyRereating}. 
In the remainder of this paper an alternative mapping approach is presented,
which generates a compact geometric track-map with a much simpler method.
It advantageously incorporates the results of our previously
published localization filter
\cite{my_references:WinterWillertEtAl2018IncreasingAccuracyTrain}, and is
furthermore more suitable for train-borne localization applications.

\section{Map Generation} % ________________________________________________
\label{sec:problem_formulation}
% • Problem formulation and the processes used to solve the problem, prove or 
%   disprove the hypothesis
% • Use illustrations to clarify ideas, support conclusions:
%     • Tables:   Present representative data or when exact values are important 
%                 to show
%     • Figures:  Quickly show ideas/conclusions that would require detailed 
%                 explanations
%     • Graphs:   Show relationships between data points or trends in data

The aim of our map generation procedure is to create compact geometric
track-maps like the example listed in \pref{tab:ref_track_map}. This table
fully represents the track shown in \pref{fig:ref_track}. The compactness
results from the fact that railway tracks always consist of a continuous
sequence of well described geometric shapes (straight, transitional arc, and
circular arc) \cite{my_references:HaldorEtAl2017Planungvonbahnanlagen}.
Therefore, a railway track can unambiguously be described by a single starting
point, the direction of the track at the starting point, the sequence of
geometric shapes, and the geometric parameters for each shape (\cf
\pref{tab:ref_track_map}).
\begin{table}[ht]
    \centering
    \caption{Compact geometric track-map for the track shown in
    \pref{fig:ref_track}.}
    \label{tab:ref_track_map}
    %
    \input{tables/ref_track_map}
\end{table}
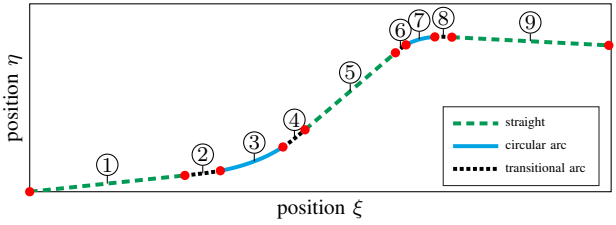
\begin{figure}[ht]
    \centering
    \tikzsetnextfilename{ref_track}
    \input{figures/ref_track}
    \caption{Exemplary track consisting of the three standard track
    geometries: straight, transitional arc and circular arc. A compact
    geometric track-map representation of this track is given in
    \pref{tab:ref_track_map}.}
    \label{fig:ref_track}
%     \vspace{-1\baselineskip}
\end{figure}

\subsection{Initial Situation}\label{sec:initial}
We assume to start with the results of the localization filter we presented in
\cite{my_references:WinterWillertEtAl2018IncreasingAccuracyTrain}. Along with
the position solution this filter estimates some of the track's geometric
parameters which conveniently serve as initialization for the map generation
process. Moreover, the filter delivers assignments between measurement
data and identified track geometries which vastly simplifies the formulation of
the mapping error that is derived later in this section.

\begin{figure}[ht]
    \centering
    \tikzsetnextfilename{initial_map}
    \input{figures/initial_map}
    \caption{Initial track-elements identified by the localization
    filter presented in
    \cite{my_references:WinterWillertEtAl2018IncreasingAccuracyTrain}. These
    elements do not constitute a continuous track-map.}
    \label{fig:initial_track_geometry}
\end{figure}
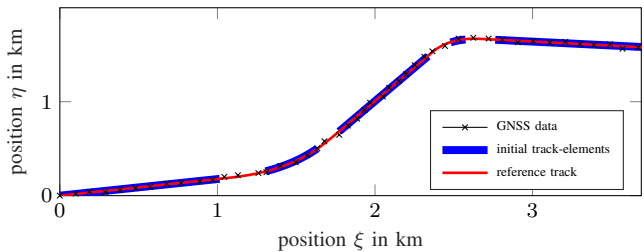

A summary of the available parameters from the localization filter is
listed in \pref{tab:initial_track_parameters}. A visualization of the resulting
discontinuous track is shown in \pref{fig:initial_track_geometry}. 
For this example, the input measurements used for the filter have been generated
by simulation. The used ground-truth track is shown in \pref{fig:ref_track}. The
detailed simulation procedure and parameters are described in
\cite{my_references:WinterWillertEtAl2018IncreasingAccuracyTrain}.
All further explanations are illustrated using this example data.
\begin{table}[ht]
    \centering
    \caption{Initially available geometric track-map parameters}
    \label{tab:initial_track_parameters}
    %
    \input{tables/initial_track_map}
\end{table}

\subsection{Mapping Procedure}
The initial track depicted in \pref{fig:initial_track_geometry} is not usable
for localization since it is discontinuous. It has some gaps at the points
where no track-geometry has been identified (\cf
\pref{tab:initial_track_parameters}, track-IDs: 2, 4, 6, and 8). Therefore, the
task of the map generation procedure is to connect the initially identified
track-elements to a continuous track which also has to fit the available GNSS
measurement data. To solve this task, first, missing track-geometries have to be
identified. Afterwards, the geometric parameters of the individual
track-elements can be estimated.

\subsubsection{Track Geometry Identification} 
We assume that the missing track-geometries can be concluded from the following
knowledge \cite{my_references:HaldorEtAl2017Planungvonbahnanlagen,
my_references:WinterWillertEtAl2018IncreasingAccuracyTrain}: Railway tracks only
consist of three basic geometric shapes which are straights, transitional arcs
and circular arcs. A straight can only be connected to a circular-arc with the
help of a transitional-arc and vice versa. Since the localization
filter already identified straights and circular-arcs it can be concluded that
the unknown geometries have to be transitional arcs.

\subsubsection{Geometry Parameter Identification} 
After all track geometries have been identified, the corresponding geometric
parameters have to be tuned such that the continuous concatenation of all
track-elements fits best to the available GNSS measurement data.
%
% \paragraph{Naive Approach}
Although there is already a lot of information available from the localization
filter, it is necessary to revise the parameters altogether.
This can be seen when we try to simply concatenate all identified tracks. The
parameters of the transitional arcs are inferred from the neighboring
elements%
\footnote{This is possible because transitional arcs are built as clothoids
\cite{my_references:HaldorEtAl2017Planungvonbahnanlagen}. They are clearly
defined by their length, their radius in the end point, and their orientation
either in the start or end-point.}.
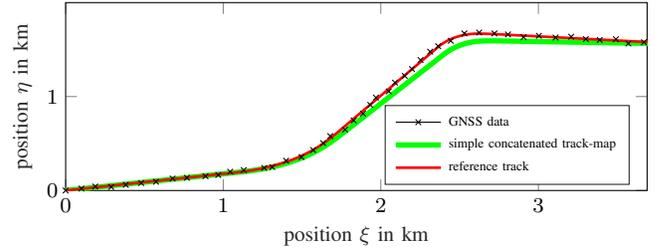
\begin{figure}[ht]
    \centering
    \tikzsetnextfilename{naive_track_map}
    \input{figures/naive_track_map}
    \caption{Track-map resulting from the simple concatenation of the 
    track-elements identified by the localization filter described in
    \cite{my_references:WinterWillertEtAl2018IncreasingAccuracyTrain} (\cf
    \pref{tab:initial_track_parameters} and \pref{fig:initial_track_geometry}).}
    \label{fig:naive_track_map}
\end{figure}
The resulting track-map for this simple approach is shown in
\pref{fig:naive_track_map}. Obviously, it is a quite poor fit to the GNSS data.
This is a result of the continuity condition of railway tracks. Slight parameter
inaccuracies of one track element are propagated on all succeeding elements.
Therefore, it is necessary to tune the parameters in a joint optimization. To
achieve this, we establish an optimization problem for the whole track by
defining an appropriate error function that incorporates all given measurement
information.
In order to solve this optimization problem we furthermore have to reformulate
the track parameters in a more suitable representation and have to choose an
optimization method. The essential aspects of the optimization are
described in the following paragraphs.

\paragraph{Error-Function Definition}
The error introduced by a track element is given by the perpendicular distances
between the track and the GNSS measurements related to this track element.
Let $\vec x_t$ be a vector representing the parameters of track element $t$.
Furthermore, let $\vec{z}_{t,i}$ be the $i$-th GNSS position measurement
assigned to this track element $t$ by the localization filter. With
$\hat{\vec{z}}_{t,i}(\vec{x}_t)$, the dropped perpendicular point of
$\vec{z}_{t,i}$ on the track element $t$, the error for this measurement is then
defined as
\begin{equation}
    \vec{e}_{t,i}(\vec{x}_t) 
    = 
    \vec{z}_{t,i} - \hat{\vec{z}}_{t,i}(\vec{x}_t)\, .
\end{equation}
The sum over all this measurement errors for all track-elements
$\mathcal T$ yields the total error of the whole track map, \ie
\begin{equation}\label{eq:error}
    F\left(\vec{\mathcal X} {=} \lbrace \vec{x}_1,\ldots,\vec{x}_N\rbrace\right)
    = 
    \sum_{t \in \mathcal T}\sum_{i \in  {\mathcal C}_t}\vec{e}_{t,i}^\Transp \vec\Omega_{t,i} \vec{e}_{t,i}\, ,
\end{equation}
where $N$ is the number of identified track-geometries, $\mathcal C_t$ is the
set of all measurements assigned to track $t$ and $\vec\Omega_{t,i}$ is the
information matrix corresponding to measurement $\vec{z}_{t,i}$. The
information matrix is the inverse of the covariance matrix which is often
provided by the GNSS receiver. If no adequate uncertainty information is
available $\vec\Omega_{t,i}$ should be chosen according to the assumed receiver
uncertainty.

\paragraph{Parameter Representation}
The parameterization of the track map presented in \pref{tab:ref_track_map} is
not very suitable for an optimization. With this parameterization the whole
track would be very sensitive to changes in specific parameters, \eg small
changes in $\varphi_0$ or $L$, would rotate, respectively move, major parts of
the track. Therefore, an alternative representation is chosen with less
sensitivity. All straights are now parameterized by their start and end point
whereas transitional arcs and circular arcs are parameterized by a minimal set
of geometric parameters. For our example track
(\cf \pref{tab:initial_track_parameters}) this new parameterization is given in
\pref{tab:reparam_track_map} and the corresponding parameter vector is
\begin{gather}
	\vec{\mathcal X} 
	= 
	\lbrace 
	\vec{x}_1, \vec{x}_2, \vec{x}_3, \vec{x}_4, \vec{x}_5, \ldots 
	\rbrace,
	\quad
	\text{with}
	\qquad\qquad\\
\begin{aligned}
%\end{equation}
% \vspace*{-2\baselineskip}
%
%\begin{align*}
    \vec{x}_1 &= \mylvec{\xi_{0,1} & \eta_{0,1} & \xi_{e,1} & \eta_{e,1}},\\
    \vec{x}_2 &= L_2,\quad
    \vec{x}_3 = \mylvec{r_3 & L_3},\quad
    \vec{x}_4 = L_4,\\
    \vec{x}_5 &= \mylvec{\xi_{0,5} & \eta_{0,5} & \xi_{e,5} & \eta_{e,5}},
    \quad\ldots\quad.
\end{aligned}\nonumber
\end{gather}

\begin{table}[ht]
    \centering
    \caption{Track-map from \pref{tab:initial_track_parameters} reparameterized
    for the optimization.}
    \label{tab:reparam_track_map}
    %
    \input{tables/reparam_track_map}
\end{table}

\paragraph{Optimization Method}
The objective is to minimize the error function
$F(\vec{\mathcal X})$ given in \pref{eq:error}. Although the initial parameters
given by the localization filter yield a poor track map when being simply
concatenated (\cf \pref{fig:naive_track_map}), they are still a good initial
guess $\vec{\mathcal X}_0$ for the track-map parameters. Therefore, we can
start the optimization with this initial parameter set which is presumably
close to the global optimum and it is sufficient to use the Levenberg-Marquardt
algorithm \cite{my_references:More1978LevenbergMarquardtalgorithm} to find that
optimum.

\section{Evaluation} % _________________________________________________________
\label{sec:analysis}
% • Describe your experiments/analysis
%     • What was the set-up, what data sources were used and why, ...
%     • What metrices are used for analysis (Why, if not already well-known)
% • Demonstrate that you solved the problem or made significant advances
% • Results: Summarized Data
%     • Should be clear and concise
%     • Use figures or tables with narrative to illustrate findings
% • Discussion: Interprets the Results
%     • Why your research offers a new solution
%     • Acknowledge any limitations
%     • use to potentially highlight connections with other work that are 
%       appreciated better after reading about the proposed work

In this section the performance of the presented mapping method will be
evaluated with the help of simulated data and real measurement data.

\subsection{Simulation Results}

First, an evaluation based on simulations is carried out to gain some
principal insights into the behavior of the presented mapping algorithm. This is
advantageous, as in simulations ground-truth data is directly available.
Throughout all simulations the example track described in \pref{sec:initial} is
used.

\subsubsection{Optimization Process}
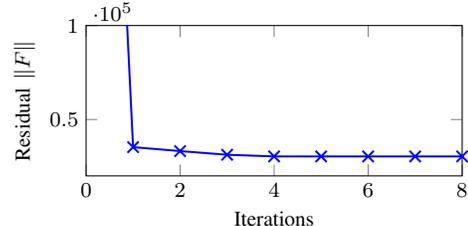
\begin{figure}[ht]
    \centering
    \tikzsetnextfilename{optim_progress}
    \input{figures/optim_progress}
    \caption{Simulation result: Progress of the residual during the
    optimization.}
    \label{fig:optim_progress}
\end{figure}

The progress of the residual $\lVert F(\vec{\mathcal X}) \rVert$ during the
optimization is shown in \pref{fig:optim_progress}. It can be seen that the
optimization converges very fast. After eight iterations the stopping criteria
(relative step size limit of \num{1e-6}) is reached. 
The biggest change in $\lVert F(\vec{\mathcal X})\rVert$ and the parameter set
occurs in the first iteration. This confirms our hypothesis that the initial
parameter set $\vec{\mathcal X}_0$ provided by the localization filter, is
already a good guess and the Levenberg-Marquardt algorithm can quickly find a
good solution.

\subsubsection{Absolute Accuracy}
A visualization of the resulting track is shown in \pref{fig:optim_map}. A good
qualitative match with the GNSS measurements and the reference track becomes
evident.
Figure \ref{fig:abs_mapping_error} allows to investigate the map's quality in
even more detail. The plot shows the absolute position deviation to the
reference track over the path length $s$. 
The generated map is compared to a typically used data-point
based map which has been sampled from the virtually generated GNSS data with a
spacing of \valunit{1}{m}. Intermediate points are calculated by a linear
interpolation.
The deviation of our optimized geometric track-map to the reference track is on
average \valunit{1.8}{m}. The error of the data-point based map varies
strongly over the whole track length and the average deviation is
\valunit{10.3}{m}. This value corresponds with the simulated GNSS measurement
noise which has a standard deviation of \valunit{10}{m}.

The better performance of the new approach results from the joint
incorporation of all measurements in the optimization process. Thereby, the
error induced by the GNSS measurement noise can be reduced significantly.

%
% \begin{table}[ht]
%     \centering
%     \caption{Final track-map parameters resulting from the optimization.}
%     \label{tab:reparam_optim_track_map}
%     %
%     \input{tables/reparam_optim_track_map}
%     %
% \end{table}
%
\begin{figure}[ht]
    \centering
    \tikzsetnextfilename{optim_map}
    \input{figures/optim_map}
    \caption{Simulation result: Final geometric track-map resulting from the
    here presented mapping approach. Due to the optimization procedure the
    final map fits very well to the GNSS data, compared to the initial map
    resulting from the simple concatenation of the identified track elements.}
    \label{fig:optim_map}
\end{figure}
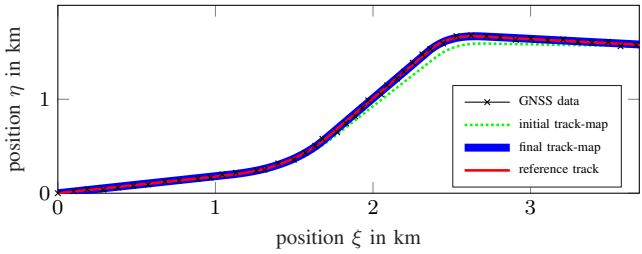
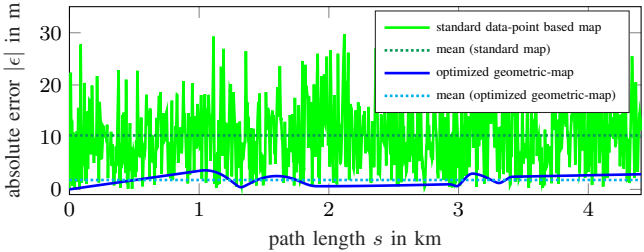
\begin{figure}[ht]
    \centering
    \tikzsetnextfilename{abs_mapping_error}
    \input{figures/abs_mapping_error}
    \caption{Simulation result: Absolute mapping error $|\epsilon|$ plotted
    against the track length $s$. The error of the geometric track-map
    generated with the here presented approach is significantly smaller
    compared to a typically used data-point based track map.}
    \label{fig:abs_mapping_error}
\end{figure}

\subsubsection{Geometric Accuracy}
An often used metric to evaluate the geometric similarity of two paths is the
fréchet distance
\cite{my_references:KubickaCelaEtAl2018ComparativeStudyApplication}. The
fréchet distance of the optimized geometric track-map to the reference track is
\valunit{3.6}{m}. In comparison, the fréchet distance of the data-point based
map is \valunit{26.9}{m}. Consequently, the generated geometric track-map is
significantly better in representing the geometric characteristic of the track.
Furthermore, the optimized geometric track-map allows to efficiently access
useful geometric information, \eg the curvature at an arbitrary path length.
For the data-point based map this is only possible with additional calculations.
%
% \begin{figure}[ht]
%     \centering
%     \tikzsetnextfilename{frechet_dist}
%     \input{figures/frechet_dist}
%     %
%     \caption{Comparison of fréchet distances to the ground-truth track. The
%     results for the optimized geometric track-map and a typically used
%     data-point based track-map are shown.}
%     \label{fig:frechet_dist}
%     %
% \end{figure}

\subsubsection{Map Size}
To compare the sizes of the maps, we express the storage demand as the number of
necessary data fields. For example, a position specification  
$
\vec{p} = \left(
        \begin{IEEEeqnarraybox*}[][c]{,c/c,}
            \xi & \eta%
        \end{IEEEeqnarraybox*}
    \right)^T
$
requires two data fields. The data-point based map which is \valunit{4.4}{km}
long and sampled at a density of \valunit{1}{m}, therefore, consist of more
than $8000$ data fields, whereas the generated track-map only consists of $38$
data fields. This clearly shows how compact the optimized geometric track-map
is, compared to a simple data-point based map.

\subsection{Evaluation on Real Measurement Data}

The mapping performance is also evaluated with real GNSS and IMU
measurement data. The data has been recorded on a \valunit{24}{km} long 
test drive on a secondary line in the Erzgebirge in Germany. The results
presented next belong to a \valunit{5.7}{km} long section of this track.

\subsubsection{Absolute Accuracy}

The absolute accuracy of the optimized geometric track-map is evaluated with the
help of OpenStreetMap (OSM) data \cite{my_references:2019opnstreetmap} since no
other reference is available.
In \pref{fig:tgc_map_result} the cumulative distribution function (CDF) of the
absolute mapping error $|\epsilon|$, \ie the
perpendicular distance between the track from the optimized geometric track-map
and the OSM map, is shown.

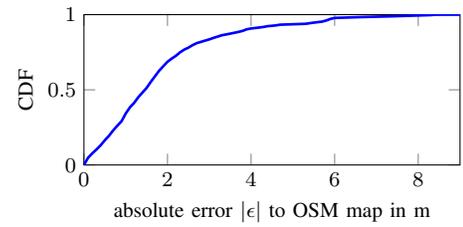
\begin{figure}[ht]
    \centering
    \tikzsetnextfilename{pd_mapping_error}
    %\tikzset{external/export next=false}  
    \input{figures/pd_mapping_error}
    \caption{Result on real measurement data: Cumulative distribution
    function (CDF) of the absolute mapping error $|\epsilon|$ between the
    optimized geometric track-map and the OSM map.}
    \label{fig:pd_mapping_error}
\end{figure}
% 
% \begin{table}[ht]
%     \centering
%     \caption{Absolute error $|\epsilon|$ in meters between the
%     generated map and the OSM map.}
%     \input{tables/track_map_quality}        
%     \label{tab:track_map_quality}
% \end{table}

The mean error is \valunit{1.8}{m} and the maximum mapping error is
\valunit{8.7}{m}. For the optimized geometric track-map a mapping error of less
than \valunit{2}{m} is achieved on \valunit{68}{\percent} of the track (\cf
\pref{fig:pd_mapping_error}). It can be concluded that the
generated map is quite accurate and that the presented mapping method is also
applicable on real data.
However, when projecting the OSM map on a satellite image, it can be seen that
the OSM map sometimes gives a poor fit to the visible course of the rails.
Interestingly, the biggest deviations between the optimized geometric track-map
and the OSM map occurs at these sections. Thus, it is very likely that the real
error of the optimized geometric track-map is even smaller than stated above.

\subsubsection{Geometric Accuracy}

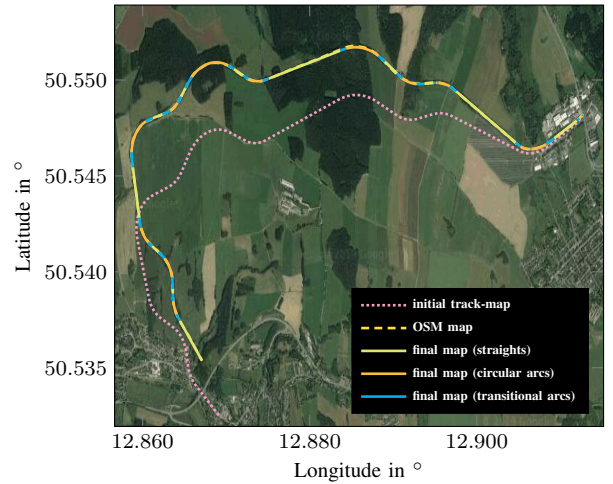
\begin{figure}[ht]
    \centering
    \tikzsetnextfilename{tgc_map_result}
    %\tikzset{external/export next=false}  
    \input{figures/tgc_map_result}
    \caption
    [
        Result on real measurement data: Visualization of the finally generated
        map on a satellite image. The OSM map is not visible on this scale as
        it is perfectly covered from the generated map.
    ]
    {
        Result on real measurement data: Visualization of the generated map on
        a satellite image\footnotemark. The OSM map is not visible on this scale
        as it is perfectly covered from the generated map.
    }
    \label{fig:tgc_map_result}
\end{figure}

The geometric accuracy of the optimized geometric track-map is evaluated with
the help of satellite images as shown in \pref{fig:tgc_map_result}.
\footnotetext{Image \textcopyright{ }2019 Google, Maps \textcopyright{ }2019
GeoBasis-DE/BKG (\textcopyright{ }2009), Google}
As in the simulational evaluation, it can be seen that the initial map,
resulting from the simple concatenation of all identified track elements,
yields a rather poor fit of the real track. Only the first track elements are
near the course of the real track (see \pref{fig:tgc_map_result} considering
the direction of travel from east to west). Due to the continuity
condition, these small errors on the first track elements accumulate for the
track elements further away of the start.

In contrast, the final map resulting from the optimization corresponds
very well to the course of the rails visible on the satellite image (\cf
\pref{fig:tgc_map_result}). Therefore, we assume that the geometric shape of the
track has been mapped in a suitable way for train-borne localization
applications. 
% An excerpt of the optimized geometric-map for the real measurement data is
% listed in \pref{tab:final_track_map}.
% 
% \begin{table}[ht]
%     \centering
%     \caption{Excerpt of the generated compact geometric track-map.}
%     \input{tables/final_track_map}        
%     \label{tab:final_track_map}
% \end{table}

\section{Conclusions} % ________________________________________________________
\label{sec:conclusions}
% • Explain what the research has achieved
%     • As it relates to the problem stated in the Introduction
%     • Revisit the key points in each section
%     • Include a summary of the main findings, important conclusions and 
%       implications for the field
% • Provide benefits and shortcomings of:
%     • The solution presented
%     • Your research and methodology
% • Suggest future areas for research

In this paper we presented an approach to generate geometric track-maps for
train-borne localization applications. After a brief overview of the
existing track-map types and map generation methods, we argued how important
track maps are for trains to become intelligent vehicles that are able to
localize themselves in the near future. Furthermore, the overview revealed that
there are hardly any adequate methods to generate suitable track maps for this
purpose.

We presented an optimization method that finds the geometric parameters of a
continuous track that fits the position measurements best and is much more
compact. The method uses information provided by a localization filter for
initialization, shape identification and data association.

Through a simulative evaluation we demonstrated that the presented method is
able to generate geometric track-maps which are more accurate than the typically
used data-point based maps. Furthermore, the generated map provides additional
geometric track information and is much more compact. Finally, we demonstrated
on a \valunit{5.7}{km} long real track-section that the approach is also
applicable on real measurement data and the resulting map corresponds very well
to the track visible in satellite images. 
Consequently, the presented method is capable of generating compact geometric
track-maps, which can help to introduce train-borne localization systems in the
near future.

\section*{Acknowledgment}
We kindly thank Deutsche Bahn for supporting this research project. Furthermore,
we like to thank Thales affording us to collect the raw data with their test
vehicle LUCY, and the group of Geodetic Measurement Systems and Sensors at TU
Darmstadt for providing the IMU/GNSS sensor platform.

\htodos{
    TODOs for possible final submission:
    \begin{itemize}
        \item check usage of hyphens in words like ``digital track-map''
        \item improve figures (maybe show some zoomed sections)
        \item consequently use track-length parameter $s$ (especially in the
        figure captions where at them moment $l$ is often used)
        \item check references and use IEEE abbreviations
        \item compare parameter tables before and after optimization
    \end{itemize}
}

%% file: tables/ref_track_map.tex
% This table requires: \usepackage{siunitx}
\begin{threeparttable}
    \renewcommand{\arraystretch}{1.3}
    \setlength{\tabcolsep}{0.5pt}
    \begin{tabular}{|r||c|c|c|c|c|c|c|c|c|} \hline
        Track ID                               & $1$         & $2$         & $3$         & $4$         & $5$         & $6$         & $7$         & $8$         & $9$         \\ \hline %\hhline{|=||=|=|=|=|=|=|=|=|=|}
        Shape\tnote{a}\hspace{0.5em}           & st          & ta          & ca          & ta          & st          &  ta         & ca          & ta          & st          \\ \hline
        Length $L$ in m                        & $1000$      & $231$       & $476$       & $231$       & $1000$      & $108$       & $206$       & $108$       & $1000$      \\ \hline
        Radius\tnote{b}\hspace{0.5em} $r$ in m & $\infty$    & $-900$      & $-900$      & $-900$      & $\infty$    & $300$       & $300$       & $300$       & $\infty$    \\ \hline
        Start point                            & \multicolumn{9}{c|}{\parbox[c][0.75cm][c]{0cm}{} $\vec{p_0} = \mylvec{0 & 0}$ \qquad $\varphi_0 = 10^\circ$}                                              \\ \hline
%         Start point $\xi_0$ in m               & $0$         & ---         & ---         & ---         & ---         & ---         & ---         & ---         & ---         \\ \hline
%         Start point $\eta_0$ in m              & $0$         & ---         & ---         & ---         & ---         & ---         & ---         & ---         & ---         \\ \hline
%         Direction $\varphi_0$ in $^\circ$      & $10$        & ---         & ---         & ---         & ---         & ---         & ---         & ---         & ---         \\ \hline
    \end{tabular}
    \begin{tablenotes}
        \item[a] st: straight,~~ta: transitional arc,~~ca: circular arc
        \item[b] the sign indicates the turning direction ($+$ right, $-$ left)
    \end{tablenotes} 
\end{threeparttable}

%% file: figures/ref_track.tex
\begin{tikzpicture}
[
baseline=(current bounding box.center)
]

\footnotesize
\begin{axis}[%
    width=7.73cm,
    height=2.5cm,
    at={(0cm,0cm)},
    scale only axis,
%     axis equal image,
    xmin=0,
    xmax=3.69,
    xlabel={position $\xi$},
    xlabel shift = -0.2cm,
    ymin=0.0,
    ymax=2.00,
    ylabel={position $\eta$},
    ylabel shift = -0.2cm,
%     axis background/.style={fill=white},
    legend style={at={(0.98,0.03)}, anchor=south east, legend cell align=left, align=left, draw=white!15!black,fill=white},
    xticklabels={,,},
    yticklabels={,,},
    xtick style={draw=none},
    ytick style={draw=none},
]
\addplot [color=ForestGreen, densely dashed, line width=1.5pt]
  table[]{figures/data/ref_track/ref_track-1.tsv};
\addlegendentry{\tiny straight}
\addplot [color=black, densely dotted, line width=1.5pt, forget plot]
  table[]{figures/data/ref_track/ref_track-2.tsv};
\addplot [color=Cerulean, solid, line width=1.5pt]
  table[]{figures/data/ref_track/ref_track-3.tsv};
\addlegendentry{\tiny circular arc}
\addplot [color=black, densely dotted, line width=1.5pt]
  table[]{figures/data/ref_track/ref_track-4.tsv};
\addlegendentry{\tiny transitional arc}
\addplot [color=ForestGreen, densely dashed, line width=1.5pt, forget plot]
  table[]{figures/data/ref_track/ref_track-5.tsv};
\addplot [color=black, densely dotted, line width=1.5pt, forget plot]
  table[]{figures/data/ref_track/ref_track-6.tsv};
\addplot [color=Cerulean, solid, line width=1.5pt, forget plot]
  table[]{figures/data/ref_track/ref_track-7.tsv};
\addplot [color=black, densely dotted, line width=1.5pt, forget plot]
  table[]{figures/data/ref_track/ref_track-8.tsv};
\addplot [color=ForestGreen, densely dashed, line width=1.5pt, forget plot]
  table[]{figures/data/ref_track/ref_track-9.tsv};
\addplot [color=red, line width=1.5pt, draw=none, mark size=1pt, mark=*, mark options={solid, red}, forget plot]
  table[]{figures/data/ref_track/ref_track-10.tsv};

\coordinate (coord_01) at (axis cs:0.4924,0.0868) {};
\coordinate (coord_02) at (axis cs:1.0983,0.1949) {};
\coordinate (coord_03) at (axis cs:1.4253,0.3234) {};
\coordinate (coord_04) at (axis cs:1.6800,0.5655) {};
\coordinate (coord_05) at (axis cs:2.0340,1.0690) {};
\coordinate (coord_06) at (axis cs:2.3525,1.5223) {};
\coordinate (coord_07) at (axis cs:2.4721,1.6216) {};
\coordinate (coord_08) at (axis cs:2.6250,1.6494) {};
\coordinate (coord_09) at (axis cs:3.1770,1.6019) {};

\node (label_01) [circle,draw=black,inner sep=0.5pt,minimum size=5pt] at ($ (coord_01) + (axis cs:+0.00,+0.20) $) {\textcolor{black}{\footnotesize $1$}}; 
\node (label_02) [circle,draw=black,inner sep=0.5pt,minimum size=5pt] at ($ (coord_02) + (axis cs:+0.00,+0.20) $) {\textcolor{black}{\footnotesize $2$}};
\node (label_03) [circle,draw=black,inner sep=0.5pt,minimum size=5pt] at ($ (coord_03) + (axis cs:+0.00,+0.20) $) {\textcolor{black}{\footnotesize $3$}};
\node (label_04) [circle,draw=black,inner sep=0.5pt,minimum size=5pt] at ($ (coord_04) + (axis cs:+0.00,+0.20) $) {\textcolor{black}{\footnotesize $4$}};
\node (label_05) [circle,draw=black,inner sep=0.5pt,minimum size=5pt] at ($ (coord_05) + (axis cs:+0.00,+0.20) $) {\textcolor{black}{\footnotesize $5$}};
\node (label_06) [circle,draw=black,inner sep=0.5pt,minimum size=5pt] at ($ (coord_06) + (axis cs:+0.00,+0.20) $) {\textcolor{black}{\footnotesize $6$}};
\node (label_07) [circle,draw=black,inner sep=0.5pt,minimum size=5pt] at ($ (coord_07) + (axis cs:+0.00,+0.20) $) {\textcolor{black}{\footnotesize $7$}};
\node (label_08) [circle,draw=black,inner sep=0.5pt,minimum size=5pt] at ($ (coord_08) + (axis cs:+0.00,+0.20) $) {\textcolor{black}{\footnotesize $8$}};
\node (label_09) [circle,draw=black,inner sep=0.5pt,minimum size=5pt] at ($ (coord_09) + (axis cs:+0.00,+0.20) $) {\textcolor{black}{\footnotesize $9$}};

\draw (coord_01) -- (label_01);
\draw (coord_02) -- (label_02);
\draw (coord_03) -- (label_03);
\draw (coord_04) -- (label_04);
\draw (coord_05) -- (label_05);
\draw (coord_06) -- (label_06);
\draw (coord_07) -- (label_07);
\draw (coord_08) -- (label_08);
\draw (coord_09) -- (label_09);
\end{axis}
\end{tikzpicture}%

%% file: tables/initial_track_map.tex
% This table requires: \usepackage{siunitx}
\begin{threeparttable}
    \renewcommand{\arraystretch}{1.3}
    \setlength{\tabcolsep}{0.5pt}
    \begin{tabular}{|r||c|c|c|c|c|c|c|c|c|} \hline
        Track ID                               & $1$         & $2$         & $3$         & $4$         & $5$         & $6$         & $7$         & $8$         & $9$         \\ \hline %\hhline{|=||=|=|=|=|=|=|=|=|=|}
        Shape\tnote{a}\hspace{0.5em}           & st          & ---         & ca          & ---         & st          & ---         & ca          & ---         & st          \\ \hline
        Length $L$ in m                        & $1035$      & $278$       & $415$       & $206$       & $983$       & $185$       & $106$       & $165$       & $962$       \\ \hline
        Radius\tnote{b}\hspace{0.5em} $r$ in m & $\infty$             & ---         & $-882$      & ---         & $\infty$    & ---         & $297$       & ---         & $\infty$    \\ \hline
        Start point $\xi_0$ in m               & $0$         & $1019$      & $1306$      & $1635$      & $1772$      & $2338$      & $2480$      & $2580$      & $2763$      \\ \hline
        Start point $\eta_0$ in m              & $0$         & $180$       & $259$       & $504$       & $683$       & $1487$      & $1631$      & $1663$      & $1660$      \\ \hline
        Direction $\varphi_0$ in $^\circ$      & $10.0$      & $10.0$      & $23.2$      & $50.1$      & $54.9$      & $54.9$      & $27.9$      & $7.4$       & $-4.8$      \\ \hline
    \end{tabular}
    \begin{tablenotes}
        \item[a] st: straight,~~ta: transitional arc,~~ca: circular arc
        \item[b] the sign indicates the turning direction ($+$ right, $-$ left)
    \end{tablenotes} 
\end{threeparttable}

%% file: figures/initial_map.tex
\begin{tikzpicture}
[
baseline=(current bounding box.center)
]

\footnotesize
\begin{axis}[%
    width=7.73cm,
    height=2.5cm,
    at={(0cm,0cm)},
    scale only axis,
%     unbounded coords=jump,
    xmin=0,
    xmax=3.69,
    xlabel style={font=\color{white!15!black}},
    xlabel={position $\xi$ in km},
    ymin=0,
    ymax=2.00,
    ylabel style={font=\color{white!15!black}},
    ylabel={position $\eta$ in km},
%     axis background/.style={fill=white},
%     axis x line*=bottom,
%     axis y line*=left,
%     xmajorgrids,
%     ymajorgrids,
    legend style={at={(0.97,0.03)}, anchor=south east, legend cell align=left, align=left, draw=white!15!black,fill=white},
    ytick={0,1},
    xtick={0,1,2,...,4},
%     xticklabels={,,},
%     yticklabels={,,},
%     xtick style={draw=none},
%     ytick style={draw=none},
]

\addplot [color=black, draw=none, mark size=1.5pt, mark=x, mark options={solid,black}] 
    table[]{figures/data/optim_map/optim_map-1.tsv};
\addlegendentry{\tiny GNSS data}

\addplot [color=blue, line width=3pt]
  table[]{figures/data/optim_map/optim_map-4.tsv};
\addlegendentry{\tiny initial track-elements}
\addplot [color=blue, line width=3pt, forget plot]
  table[]{figures/data/optim_map/optim_map-5.tsv};
\addplot [color=blue, line width=3pt, forget plot]
  table[]{figures/data/optim_map/optim_map-6.tsv};
\addplot [color=blue, line width=3pt, forget plot]
  table[]{figures/data/optim_map/optim_map-7.tsv};
\addplot [color=blue, line width=3pt, forget plot]
  table[]{figures/data/optim_map/optim_map-8.tsv};
  
  \addplot [color=red, line width=1pt]
  table[]{figures/data/optim_map/optim_map-2.tsv};
\addlegendentry{\tiny reference track}
  
% \addplot [color=blue, dotted, line width=0.75pt]
%   table[]{figures/data/optim_map/optim_map-3.tsv};
% \addlegendentry{\tiny initial cont. map}

% \addplot [color=green, line width=0.75pt]
%   table[]{figures/data/optim_map/optim_map-9.tsv};
% \addlegendentry{\tiny lin. interpol. map}

% \addplot [color=blue, line width=0.75pt]
%   table[]{figures/data/optim_map/optim_map-10.tsv};
% \addlegendentry{\tiny optim. map}

\end{axis}
\end{tikzpicture}%

%% file: figures/naive_track_map.tex
\begin{tikzpicture}
[
baseline=(current bounding box.center)
]

\footnotesize
\begin{axis}[%
    width=7.73cm,
    height=2.5cm,
    at={(0cm,0cm)},
    scale only axis,
%     unbounded coords=jump,
    xmin=0,
    xmax=3.69,
    xlabel style={font=\color{white!15!black}},
    xlabel={position $\xi$ in km},
    ymin=0,
    ymax=2.00,
    ylabel style={font=\color{white!15!black}},
    ylabel={position $\eta$ in km},
%     axis background/.style={fill=white},
%     axis x line*=bottom,
%     axis y line*=left,
%     xmajorgrids,
%     ymajorgrids,
    legend style={at={(0.97,0.03)}, anchor=south east, legend cell align=left, align=left, draw=white!15!black,fill=white},
    ytick={0,1},
    xtick={0,1,2,...,4},
%     xticklabels={,,},
%     yticklabels={,,},
%     xtick style={draw=none},
%     ytick style={draw=none},
]

\addplot [color=black, draw=none, mark size=1.5pt, mark=x, mark options={solid,black}]
  table[]{figures/data/optim_map/optim_map-1.tsv};
\addlegendentry{\tiny GNSS data}

% \addplot [color=blue, line width=0.75pt]
%   table[]{figures/data/optim_map/optim_map-4.tsv};
% \addlegendentry{\tiny initial track-elements}
% \addplot [color=blue, line width=0.75pt, forget plot]
%   table[]{figures/data/optim_map/optim_map-5.tsv};
% \addplot [color=blue, line width=0.75pt, forget plot]
%   table[]{figures/data/optim_map/optim_map-6.tsv};
% \addplot [color=blue, line width=0.75pt, forget plot]
%   table[]{figures/data/optim_map/optim_map-7.tsv};
% \addplot [color=blue, line width=0.75pt, forget plot]
%   table[]{figures/data/optim_map/optim_map-8.tsv};
  
\addplot [color=green, line width=2pt]
  table[]{figures/data/optim_map/optim_map-3.tsv};
\addlegendentry{\tiny simple concatenated track-map}

\addplot [color=red, line width=1pt]
  table[]{figures/data/optim_map/optim_map-2.tsv};
\addlegendentry{\tiny reference track}

% \addplot [color=green, line width=0.75pt]
%   table[]{figures/data/optim_map/optim_map-9.tsv};
% \addlegendentry{\tiny lin. interpol. map}

% \addplot [color=blue, line width=0.75pt]
%   table[]{figures/data/optim_map/optim_map-10.tsv};
% \addlegendentry{\tiny optim. map}

\end{axis}
\end{tikzpicture}%

%% file: tables/reparam_track_map.tex
% This table requires: \usepackage{siunitx}
\begin{threeparttable}
    \renewcommand{\arraystretch}{1.3}
    \setlength{\tabcolsep}{0.5pt}
    \begin{tabular}{|r||c|c|c|c|c|c|c|c|c|} \hline
        Track ID                               & $1$         & $2$         & $3$         & $4$         & $5$         & $6$         & $7$         & $8$         & $9$         \\ \hline %\hhline{|=||=|=|=|=|=|=|=|=|=|}
        Shape\tnote{a}\hspace{0.5em}           & st          & ta           & ca         & ta          & st          & ta          & ca          & ta          & st          \\ \hline
        Length $L$ in m                        & ---         & $278$       & $415$       & $206$       & ---         & $185$       & $106$       & $165$       & ---         \\ \hline
        Radius\tnote{b}\hspace{0.5em} $r$ in m & ---         & ---         & $-882$      & ---         & ---         & ---         & $297$       & ---         & ---         \\ \hline
        Start point $\xi_0$ in m               & $0$         & ---         & ---         & ---         & $1772$      & ---         & ---         & ---         & $2763$      \\ \hline
        Start point $\eta_0$ in m              & $0$         & ---         & ---         & ---         & $683$       & ---         & ---         & ---         & $1660$      \\ \hline
        End point $\xi_e$ in m                 & $1019$      & ---         & ---         & ---         & $2338$      & ---         & ---         & ---         & $3722$      \\ \hline
        End point $\eta_e$ in m                & $180$       & ---         & ---         & ---         & $1487$      & ---         & ---         & ---         & $1579$      \\ \hline
    \end{tabular}
    \begin{tablenotes}
        \item[a] st: straight,~~ta: transitional arc,~~ca: circular arc
        \item[b] the sign indicates the turning direction ($+$ right, $-$ left)
    \end{tablenotes} 
\end{threeparttable}

%% file: figures/optim_progress.tex
\begin{tikzpicture}
[
baseline=(current bounding box.center)
]

\footnotesize
\begin{axis}[%
    width=5cm,
    height=2cm,
    at={(0cm,0cm)},
    scale only axis,
    xmin=0,
    xmax=8,
    xlabel={Iterations},
    ymin=20000,
%     ymin=3.035e+04,
    ymax=100000,
    ytick={0.5e+5,1e+5},
%     ymax=3.800e+04, % 5.458e+05
%     ylabel style={font=\color{white!15!black}},
    ylabel={Residual $\lVert F\rVert$},
%     axis background/.style={fill=white},
%     axis x line*=bottom,
%     axis y line*=left,
%     xmajorgrids,
%     ymajorgrids,
%     legend style={legend cell align=left, align=left, draw=white!15!black}
]
\addplot [color=blue, line width=0.75pt, mark size=3pt, mark=x, mark
options={solid, blue}] table[]{figures/data/optim_progress/optim_progress-1.tsv};
% \addlegendentry{\tiny GPS}

\end{axis}
\end{tikzpicture}%

%% file: figures/optim_map.tex
\begin{tikzpicture}
[
baseline=(current bounding box.center)
]

\footnotesize
\begin{axis}[%
    width=7.73cm,
    height=2.5cm,
    at={(0cm,0cm)},
    scale only axis,
%     unbounded coords=jump,
    xmin=0,
    xmax=3.69,
    xlabel style={font=\color{white!15!black}},
    xlabel={position $\xi$ in km},
    ymin=0,
    ymax=2.00,
    ylabel style={font=\color{white!15!black}},
    ylabel={position $\eta$ in km},
%     axis background/.style={fill=white},
%     axis x line*=bottom,
%     axis y line*=left,
%     xmajorgrids,
%     ymajorgrids,
    legend style={at={(0.97,0.03)}, anchor=south east, legend cell align=left, align=left, draw=white!15!black,fill=white},
    ytick={0,1},
    xtick={0,1,2,...,4},
%     xticklabels={,,},
%     yticklabels={,,},
%     xtick style={draw=none},
%     ytick style={draw=none},
]

\addplot [color=black, draw=none, mark size=1.5pt, mark=x, mark options={solid,black}]
  table[]{figures/data/optim_map/optim_map-1.tsv};
\addlegendentry{\tiny GNSS data}

\addplot [color=green, densely dotted, line width=1pt]
  table[]{figures/data/optim_map/optim_map-3.tsv};
\addlegendentry{\tiny initial track-map}

% \addplot [color=cyan, line width=1.5pt]
%   table[]{figures/data/optim_map/optim_map-4.tsv};
% \addlegendentry{\tiny initial track-elements}
% \addplot [color=cyan, line width=1.5pt, forget plot]
%   table[]{figures/data/optim_map/optim_map-5.tsv};
% \addplot [color=cyan, line width=1.5pt, forget plot]
%   table[]{figures/data/optim_map/optim_map-6.tsv};
% \addplot [color=cyan, line width=1.5pt, forget plot]
%   table[]{figures/data/optim_map/optim_map-7.tsv};
% \addplot [color=cyan, line width=1.5pt, forget plot]
%   table[]{figures/data/optim_map/optim_map-8.tsv};

% \addplot [color=green, line width=0.75pt]
%   table[]{figures/data/optim_map/optim_map-9.tsv};
% \addlegendentry{\tiny lin. interpol. map}

\addplot [color=blue, line width=3pt]
  table[]{figures/data/optim_map/optim_map-10.tsv};
\addlegendentry{\tiny final track-map}

\addplot [color=red, line width=1pt]
  table[]{figures/data/optim_map/optim_map-2.tsv};
\addlegendentry{\tiny reference track}

\end{axis}
\end{tikzpicture}%

%% file: figures/abs_mapping_error.tex
\begin{tikzpicture}
[
baseline=(current bounding box.center)
]

\footnotesize
\begin{axis}[%
    width=7.60cm,
    height=2.5cm,
    at={(0cm,0cm)},
    scale only axis,
    xmin=0,
    xmax=4.407,
    xlabel style={font=\color{white!15!black}},
    xlabel={path length $s$ in km},
    ymin=-1,
    ymax=35,
    ylabel style={font=\color{white!15!black}},
    ylabel={absolute error $|\epsilon|$ in m},
%     axis background/.style={fill=white},
    xtick={0,1,2,...,4},
    ytick={0,10,...,50},
%     axis x line*=bottom,
%     axis y line*=left,
    legend style={legend cell align=left, align=left, draw=white!15!black,fill=white}
]
\addplot [color=green, line width=1.0pt]
  table[]{figures/data/abs_mapping_error/abs_mapping_error-1.tsv};
\addlegendentry{\tiny standard data-point based map}

\addplot [color=ForestGreen, densely dotted, line width=1.0pt]
  table[]{figures/data/abs_mapping_error/abs_mapping_error-2.tsv};
\addlegendentry{\tiny mean (standard map)}

\addplot [color=blue, line width=1.0pt]
  table[]{figures/data/abs_mapping_error/abs_mapping_error-3.tsv};
\addlegendentry{\tiny optimized geometric-map}

\addplot [color=Cyan, densely dotted, line width=1.0pt]
  table[]{figures/data/abs_mapping_error/abs_mapping_error-4.tsv};
\addlegendentry{\tiny mean (optimized geometric-map)}

\end{axis}
\end{tikzpicture}%

%% file: figures/pd_mapping_error.tex
\footnotesize
\begin{tikzpicture}

\begin{axis}[%
width=5cm,
height=2cm,
at={(0cm,0cm)},
scale only axis,
xmin=0,
xmax=9,
xlabel={absolute error $|\epsilon|$ to OSM map in m},
ymin=0,
ymax=1,
ylabel={CDF},
ytick={0,0.5,1},
axis background/.style={fill=white},
% axis x line*=bottom,
% axis y line*=left,
% legend style={legend cell align=left, align=left, draw=white!15!black}
]
\addplot [color=blue, line width=1.0pt]
  table[]{figures/data/pd_mapping_error/pd_mapping_error-1.tsv};
% \addlegendentry{data1}

\end{axis}
\end{tikzpicture}%

%% file: figures/tgc_map_result.tex
\definecolor{mycolor1}{rgb}{1.00000,0.00000,1.00000}%
\definecolor{mycolor2}{rgb}{0.00000,1.00000,1.00000}%
\definecolor{mycolor3}{rgb}{1.00000,1.00000,0.00000}%
\footnotesize
\begin{tikzpicture}

\begin{axis}[%
width=6.5cm,
at={(0cm,0cm)},
scale only axis,
point meta min=0,
point meta max=1,
axis on top,
unbounded coords=jump,
xmin=12.856643,
xmax=12.915178,
xlabel={Longitude in $^{\circ}$ },
ymin=50.531957,
ymax=50.55392,
ylabel={Latitude in $^{\circ}$},
% axis background/.style={fill=white},
% axis x line*=bottom,
% axis y line*=left,
% xmajorgrids,
% ymajorgrids,
max space between ticks=100pt,
tick label style={/pgf/number format/.cd,fixed,fixed zerofill,precision=3},
legend style={at={(0.97,0.03)}, anchor=south east, legend cell align=left,align=left, fill=black}
]
\addplot [forget plot] graphics [xmin=12.831022, xmax=12.940885, ymin=50.577862, ymax=50.508044] {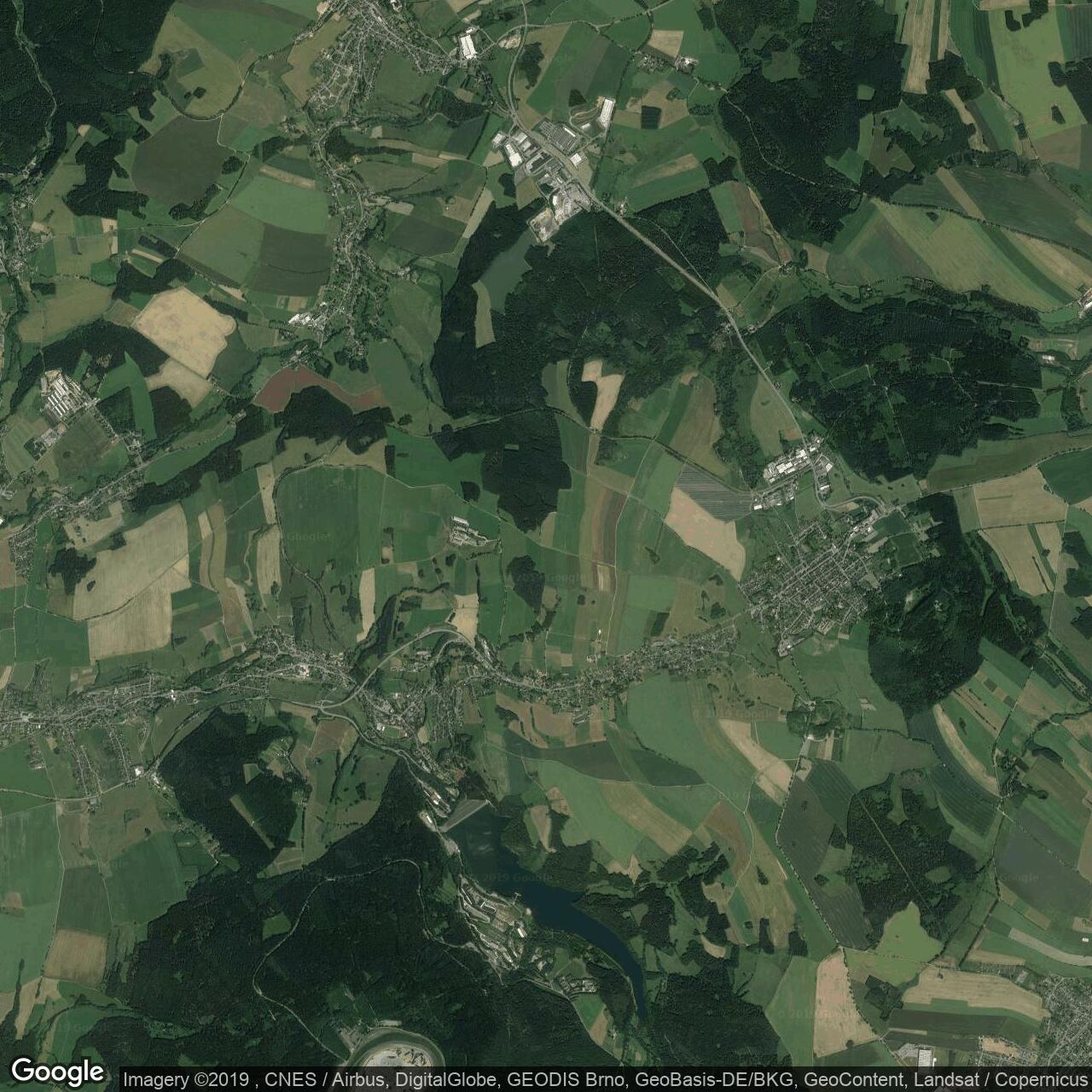};
\addplot [color=Lavender, line width=1.0pt, densely dotted]
  table[]{figures/data/tgc_map_result/tgc_map_result-1.tsv};
\addlegendentry{\textbf{\tiny\textcolor{white}{initial track-map}}}

% \addplot [color=lightgray, line width=4pt]
%   table[]{figures/data/tgc_map_result/tgc_map_result-2.tsv};
% \addlegendentry{\tiny\textcolor{white}{TGC Final Map}}

\addplot [color=Goldenrod, line width=1.0pt,densely dashed]
  table[]{figures/data/tgc_map_result/tgc_map_result-8.tsv};
\addlegendentry{\textbf{\tiny\textcolor{white}{OSM map}}}

\addplot [color=GreenYellow, line width=1.0pt]
  table[]{figures/data/tgc_map_result/tgc_map_result-3.tsv};
\addlegendentry{\textbf{\tiny\textcolor{white}{final map (straights)}}}

\addplot [color=Dandelion, line width=1.0pt]
  table[]{figures/data/tgc_map_result/tgc_map_result-4.tsv};
\addlegendentry{\textbf{\tiny\textcolor{white}{final map (circular arcs)}}}

\addplot [color=Cyan, line width=1.0pt]
  table[]{figures/data/tgc_map_result/tgc_map_result-5.tsv};
\addlegendentry{\textbf{\tiny\textcolor{white}{final map (transitional arcs)}}}

% \addplot [color=blue, line width=1.5pt]
%   table[]{figures/data/tgc_map_result/tgc_map_result-6.tsv};
% \addlegendentry{TGC Straight Tracks}

% \addplot [color=red, line width=1.5pt]
%   table[]{figures/data/tgc_map_result/tgc_map_result-7.tsv};
% \addlegendentry{TGC Circular Tracks}

\end{axis}
\end{tikzpicture}%